\documentclass[letterpaper, 10 pt, conference]{ieeeconf}  

\IEEEoverridecommandlockouts                              

\makeatletter
\let\NAT@parse\undefined
\makeatother

\usepackage{cite}
\usepackage{amsmath,amssymb,amsfonts}
\usepackage{algorithmic}
\usepackage{graphicx}
\usepackage{pifont}
\usepackage{textcomp}
\usepackage{hyperref}
\usepackage{multirow}
\usepackage{balance}
\usepackage[table,xcdraw]{xcolor}
\usepackage{url}
\usepackage{comment}
\usepackage{color}
\usepackage{latexsym}
\usepackage{mathtools}
\usepackage{caption}
\usepackage{subcaption}
\usepackage{orcidlink}

\title{
Metaverse for Safer Roadways: An Immersive Digital Twin Framework for Exploring Human-Autonomy Coexistence in Urban Transportation Systems
}
\author{Tanmay V. Samak$^{\ast}$ \orcidlink{0000-0002-9717-0764}, Chinmay V. Samak$^{\ast}$ \orcidlink{0000-0002-6455-6716} and Venkat N. Krovi\orcidlink{0000-0003-2539-896X}
\thanks{$^{\ast}$These authors contributed equally.}
\thanks{T. V. Samak, C. V. Samak, and V. N. Krovi are with the Department of Automotive Engineering, Clemson University International Center for Automotive Research (CU-ICAR), Greenville, SC 29607, USA. Email: {\tt\small {\{\href{mailto:tsamak@clemson.edu}{tsamak}, \href{mailto:csamak@clemson.edu}{csamak}, \href{mailto:vkrovi@clemson.edu}{vkrovi}\}@clemson.edu}}}
}

\begin{document}

\maketitle
\thispagestyle{empty}
\pagestyle{empty}


\begin{abstract}
Societal-scale deployment of autonomous vehicles requires them to coexist with human drivers, necessitating mutual understanding and coordination among these entities. However, purely real-world or simulation-based experiments cannot be employed to explore such complex interactions due to safety and reliability concerns, respectively. Consequently, this work presents an immersive digital twin framework to explore and experiment with the interaction dynamics between autonomous and non-autonomous traffic participants. Particularly, we employ a mixed-reality human-machine interface to allow human drivers and autonomous agents to observe and interact with each other for testing edge-case scenarios while ensuring safety at all times. To validate the versatility of the proposed framework's modular architecture, we first present a discussion on a set of user experience experiments encompassing 4 different levels of immersion with 4 distinct user interfaces. We then present a case study of uncontrolled intersection traversal to demonstrate the efficacy of the proposed framework in validating the interactions of a primary human-driven, autonomous, and connected autonomous vehicle with a secondary semi-autonomous vehicle. The proposed framework has been openly released to guide the future of autonomy-oriented digital twins and research on human-autonomy coexistence.\\%
\end{abstract}



\section{Introduction}
\label{Section: Introduction}

Autonomous vehicles cannot be deployed overnight, nor can human drivers be expected to stop driving their vehicles abruptly. Consequently, the two entities will have to co-exist on the roads, which calls for them to comprehend each other and cooperate together. This calls for imparting autonomous vehicles with ``social awareness'' towards their peers \cite{loke2019, crosato2024, phani2024}, which includes other autonomous vehicles as well as human drivers. Particularly, autonomous agents will have to learn to predict the behavior of their peers and coordinate with them in order to achieve the collective goal of safe and efficient navigation.

Today, we see that most of the robotaxi companies have matured their prototype autonomous vehicles to operate in complex scenes and scenarios. However, one of the major challenges faced by the industry today is testing their vehicles on live streets \cite{zheng2023, cadmv2024, nhtsa2023, ntsb2024}. It is becoming increasingly difficult to trace the liability of accidents caused by autonomous vehicles \cite{joseph2021}, especially those also involving human drivers.

\begin{figure}[t]
     \centering
     \includegraphics[width=\linewidth]{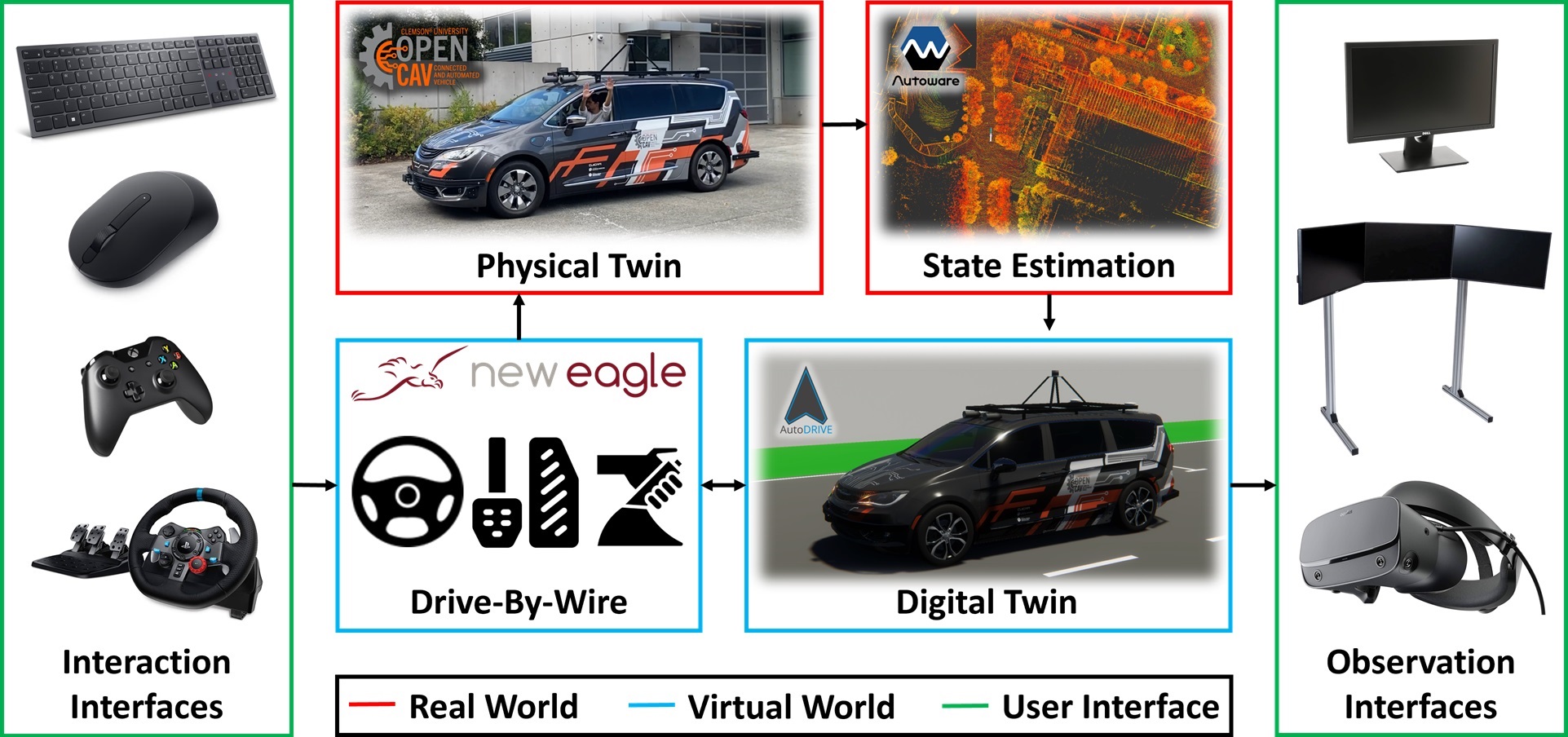}
     \caption{Simplified schematic of the immersive digital twin framework for exploring human-autonomy coexistence.}
     \label{fig1}
\end{figure}

To this end, it seems reasonable to focus AV testing efforts towards socially aware autonomous driving. However, purely real-world or simulation-based experiments cannot be employed to explore such complex interaction dynamics due to safety and reliability concerns, respectively. While real-world testing can capture the variability in terms of social behaviors to a reasonable extent, it can be potentially hazardous and unethical. Simulation-based testing, on the other hand, can provide a safe virtual proving ground to explore social interactions, but the fidelity of such interactions is often limited. Simulated traffic participants are often modeled as non-playable characters (NPCs) with heuristic behavior models \cite{CARLA2017, LGSVLSimulator2020}, data-driven social models \cite{wang2021, chen2017}, or stochastic micro-simulations \cite{SUMO2018}. However, human drivers have distinct behaviors ranging from over-defensive to over-aggressive driving, which typically varies depending on the other traffic participants and situations. Thus, human-in-the-loop presence \cite{arppe2020, lauren2023, helen2024} is needed, along with a supporting immersive digital twin framework for safely exploring human-autonomy coexistence.

In this paper, we present AutoDRIVE Ecosystem\footnote{\textbf{Website:} \url{https://autodrive-ecosystem.github.io}} \cite{AutoDRIVE2023, AutoDRIVEReport2022} as an immersive digital twin framework to bridge the gap between real and virtual worlds, facilitating a safe and reliable platform for exploring human-autonomy coexistence. The proposed framework is capable of immersing human drivers in the loop with hybrid traffic participants such that they can observe and interact with each other in real-time. This way, there is a two-fold benefit: (a) autonomous vehicles can be assessed and improved with human-in-the-loop testing, and (b) humans can become more familiar and comfortable with autonomous vehicles and both traffic entities can harmoniously coexist on the roads.

\begin{figure*}[t]
     \centering
     \begin{subfigure}[b]{0.49\linewidth}
         \centering
         \includegraphics[width=\linewidth]{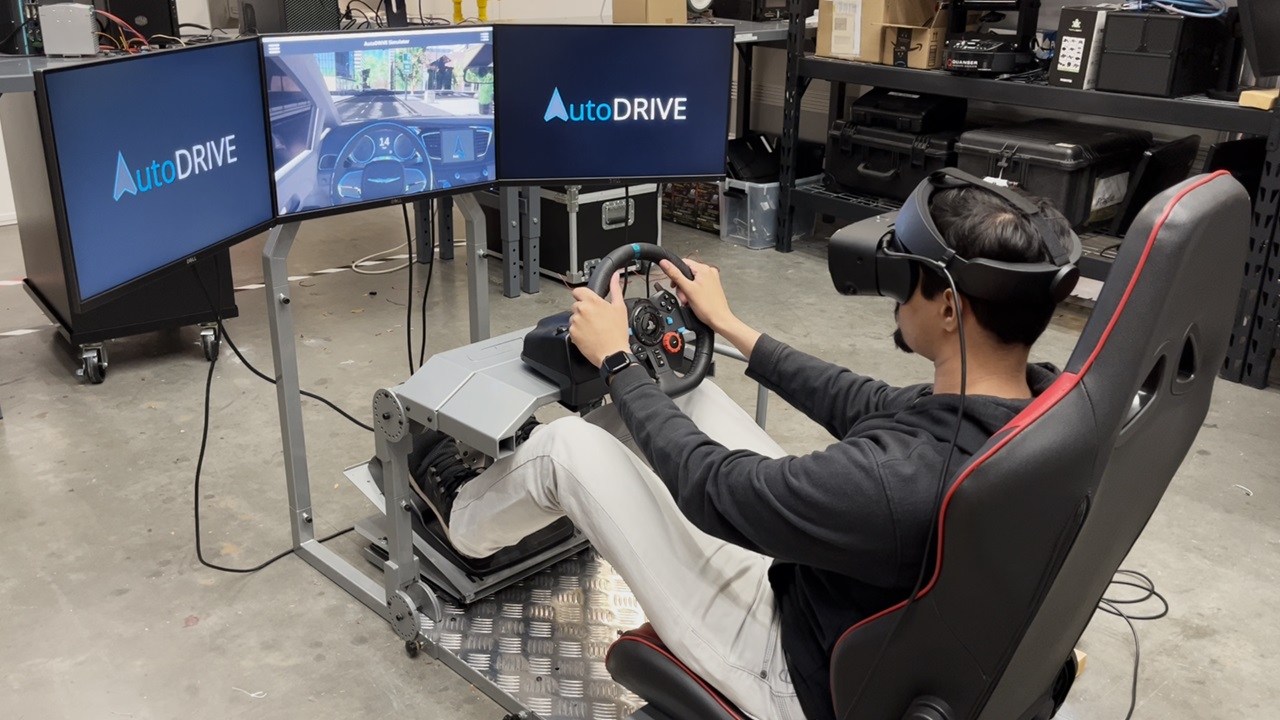}
         \caption{Exerimental setup for the in-lab user study.}
         \label{fig2a}
     \end{subfigure}
     \hfill
     \begin{subfigure}[b]{0.49\linewidth}
         \centering
         \includegraphics[width=\linewidth]{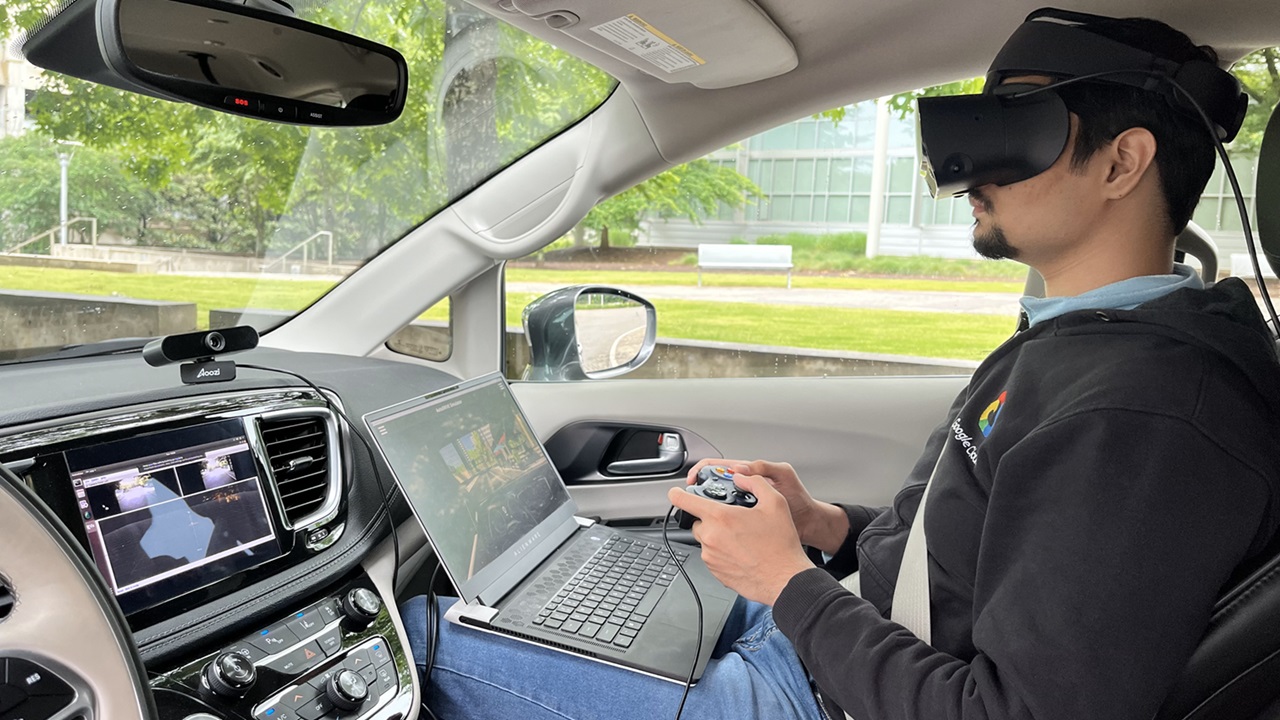}
         \caption{Exerimental setup for the on-campus case study.}
         \label{fig2b}
     \end{subfigure}
     \caption{Experimental methodology for validating the user experience and serviceability of the proposed framework.}
    \label{fig2}
\end{figure*}

The key contributions of this paper can be summarized as follows:
\begin{itemize}
    \item Development of a digital twin framework, which connects the real and virtual worlds in real-time, and can immerse humans with the physical and digital assets using various observation and interaction interfaces.
    \item A mixed-factorial user study evaluating the effectiveness of various observation and interaction interfaces to the proposed framework.
    \item A multi-factor case study comparing the interactions of a primary human-driven, autonomous, and connected autonomous vehicle with a secondary semi-autonomous vehicle.
\end{itemize}

The remainder of this paper is organized as follows. Section \ref{Section: Materials and Methods} elucidates the proposed immersive digital twin framework and research methodology adopted along with the survey and research instruments adopted. Section \ref{Section: User Study} presents the user experience study highlighting the participant demographics, design of experiments, and results. Section \ref{Section: Case Study} discusses the mixed human-autonomy ``jump scare'' edge-case study highlighting the driving scenario, design of experiments, and results. Finally, Section \ref{Section: Conclusion} summarizes the overall findings of this work and highlights some of the potential future directions of this research.


\section{Materials and Methods}
\label{Section: Materials and Methods}

\subsection{Digital Twin Framework}
\label{Section: Digital Twin Framework}

The proposed immersive digital twin framework (refer Fig. \ref{fig1}) is composed of 4 major components: (a) physical twin of the ego traffic participant; (b) digital twin(s) of the ego and non-ego traffic participant(s); (c) observation modalities for the ego traffic participant; and (d) interaction modalities for the ego traffic participant. Apart from this, there exist supporting elements such as state estimation of the physical twin to update the digital twin; simulating virtual environment and infrastructure, sensor physics, vehicle dynamics, and traffic flow; observing, interacting, and decision-making using the digital twin with or without a human driver in the loop; and passing the control actions to the physical twin in real-time to complete the digital thread.

Being capable of seamlessly bridging the real2sim \cite{Real2Sim2Real2024} and sim2real \cite{samak2023sim2real} gaps, the proposed digital twin framework facilitates core components for vehicle autonomy and smart-city infrastructure and integrates open-interface libraries, plugins, tools, and application programming interfaces (APIs) to accelerate the development and validation of modern intelligent transportation systems (ITS). The key features of the proposed framework stand out in the following aspects:

\begin{itemize}
    \item \textbf{Digital Twinning:} Simulating physically and graphically accurate and calibrated vehicles with real-time, bi-directional updates between the physical and digital realms.
    \item \textbf{Mixed Reality:} Selectively observing/interacting with real and/or virtual environments (terrain, sky, time, weather, etc.), infrastructure (roads, buildings, traffic signs, traffic lights, etc.), and traffic (human drivers, heuristic models, social models, stochastic micro-simulations, etc.)
    \item \textbf{Human-in-the-Loop:} Immersing humans in the loop with autonomous and non-autonomous vehicles for devising socially-aware autonomy algorithms and studying human factors in ITS.
    \item \textbf{V2X Communication:} Vehicle-to-vehicle (V2V), vehicle-to-infrastructure (V2I) and vehicle-to-human (V2H) communication between real/virtual traffic participants with the possibility of integrating network simulator such as \cite{Riley2010}.
    \item \textbf{Platform-Agnostic:} Cross-platform toolchain for small, mid, and full-scale vehicles targeting diverse operational design domains (ODDs) and applications \cite{DTAcrossScales2024}.
    \item \textbf{Modular:} Modular and reconfigurable architecture with open interfaces to \textit{``plug and play''} with third-party hardware/software plugins and expand the framework.
    \item \textbf{Open-Source:} We release the presented immersive digital twin framework as an open-source\footnote{\textbf{GitHub:} \url{https://github.com/autodrive-ecosystem}} effort and hope that its modular architecture can guide the future development of digital twins and research pertaining to human-autonomy coexistence.
\end{itemize}

\begin{figure*}[t]
     \centering
     \includegraphics[width=\linewidth]{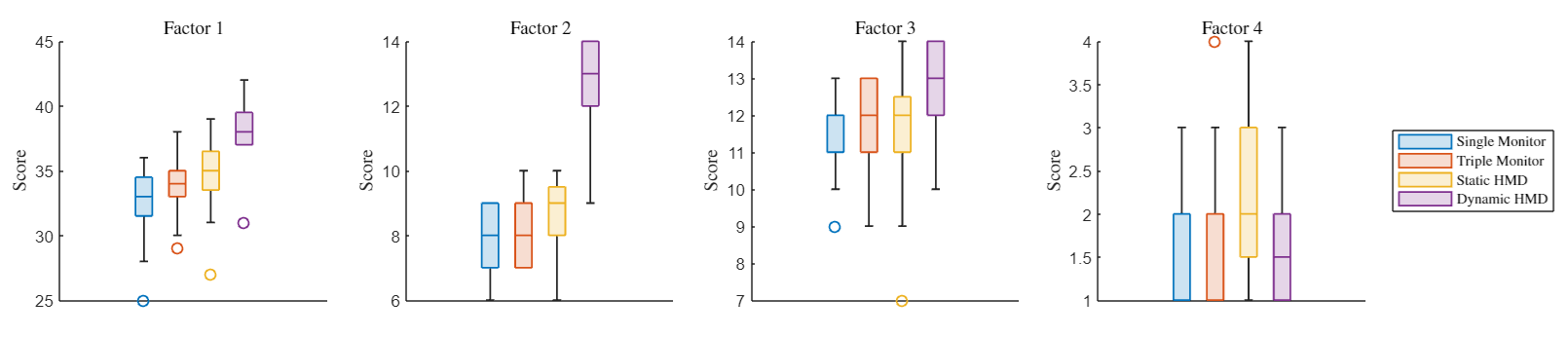}
     \caption{User experience analysis for observation interfaces: Scoring for the 4-factor model of the 11-item PQ. Factor 1 is scored out of 42, factors 2 and 3 are scored out of 14, while factor 4 is scored out of 7, based on the number of questions per factor.}
     \label{fig3}
\end{figure*}

\begin{figure*}[t]
     \centering
     \includegraphics[width=\linewidth]{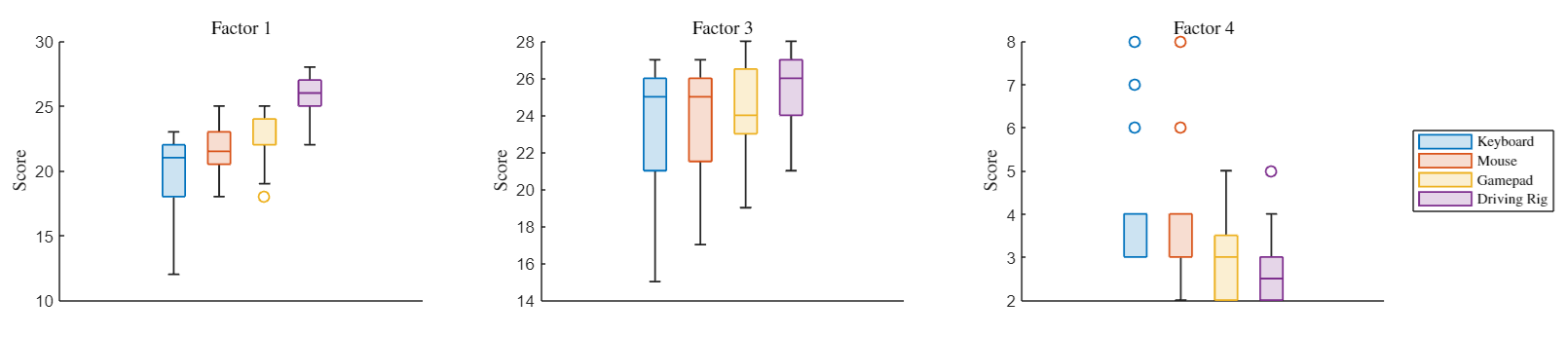}
     \caption{User experience analysis for interaction interfaces: Scoring for the 3-factor model (factor 2 is not relevant to the interaction interface) of the 10-item PQ. Factors 1 and 3 are scored out of 28, while factor 4 is scored out of 14, based on the number of questions per factor.}
     \label{fig4}
\end{figure*}

\subsection{Research Instrument}
\label{Section: Research Instrument}

In this work, the ego traffic participant of choice was the Open Connected and Automated Vehicle (OpenCAV)\footnote{\textbf{Website:} \url{https://sites.google.com/view/opencav}} housed at CU-ICAR. The OpenCAV is a 2018 Chrysler Pacifica, retrofitted with a modular sensor suite for interoceptive and exteroceptive perception. These include a standalone, dual-antenna GNSS (Novatel PwrPak7D) and MEMS-grade inertial measurement unit (Novatel IMU-IGM-S1), 2 HD cameras (Mako G-319) with distinct focal lengths (right 12 mm and left 16 mm), an electronically scanning RADAR (Aptiv ESR 2.5), and a 32-beam 3D LIDAR (Velodyne VLP-32C). The vehicle utilizes a robust GPU computing edge AI platform (AStuff Spectra ECU with Intel Xeon CPU and NVIDIA RTX 2080 GPU) for onboard computation and houses a 23 TB onboard data storage equipment (Quantum R3000). Connectivity to the outside world is facilitated through a dedicated Cradlepoint 5G wireless network router (CPI-IBR900-US). Vehicle telemetry data is easily accessible via the USB-CAN bus (Kvaser USBcan 4xHS) connected to the ECU. The drive-by-wire system (New Eagle ECM196) of the vehicle is capable of controlling the steering, throttle, brake, and gear shift, along with auxiliary functions like lights, doors, and wipers. The software stack of OpenCAV is developed atop Autoware, which facilitates continuous development and integration of autonomy modules for mapping, perception, localization, motion planning and control.\\

We created an accurate digital twin representation of OpenCAV within AutoDRIVE Simulator \cite{AutoDRIVESimulator2021, AutoDRIVESimulatorReport2020} by calibrating it against its physical counterpart. This digital twin can be plugged into any virtual environment on demand and simulated independently or co-simulated with a third-party program, or digitally twinned in the loop with the physical vehicle. From a computational perspective, the said simulator was developed modularly using object-oriented programming (OOP) constructs and leveraged CPU multi-threading as well as GPU instancing (if available) to efficiently parallelize various simulation objects and processes. The simulator facilitates open APIs for C\#, C++, Python, MATLAB, ROS, ROS 2, Autoware, and Webapp. For this work, all the simulations were executed on a laptop PC with Intel Core i9-12900H CPU, NVIDIA RTX 3080 Ti GPU and 32 GB RAM.

In order to facilitate observation and interaction with the physical/digital assets, we integrated various human-machine interfaces (HMIs) with the proposed digital twin framework. In terms of observation interfaces, this study utilized a single monitor, a triple monitor setup, and a head-mounted display (HMD), particularly the Oculus Rift S. In terms of interaction interfaces, this study utilized a standard keyboard, an optical mouse, a gamepad (Xbox Wireless Controller), and a driving rig (Logitech G29). The input/output mappings of all the modalities were calibrated to perceive and control vehicles of different scales and configurations\footnote{\textbf{Video:} \url{https://youtu.be/_cwrw1w5d_g}}.

\begin{figure*}[t]
     \centering
     \begin{subfigure}[b]{0.49\linewidth}
         \centering
         \includegraphics[width=\linewidth]{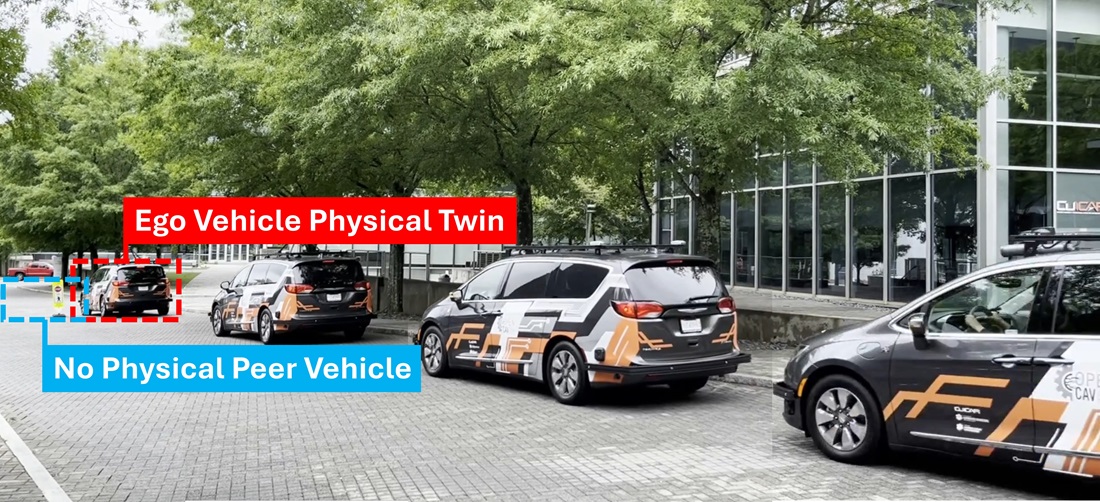}
         \caption{Ego vehicle physical twin with no traffic participant(s).}
         \label{fig5a}
     \end{subfigure}
     \hfill
     \begin{subfigure}[b]{0.49\linewidth}
         \centering
         \includegraphics[width=\linewidth]{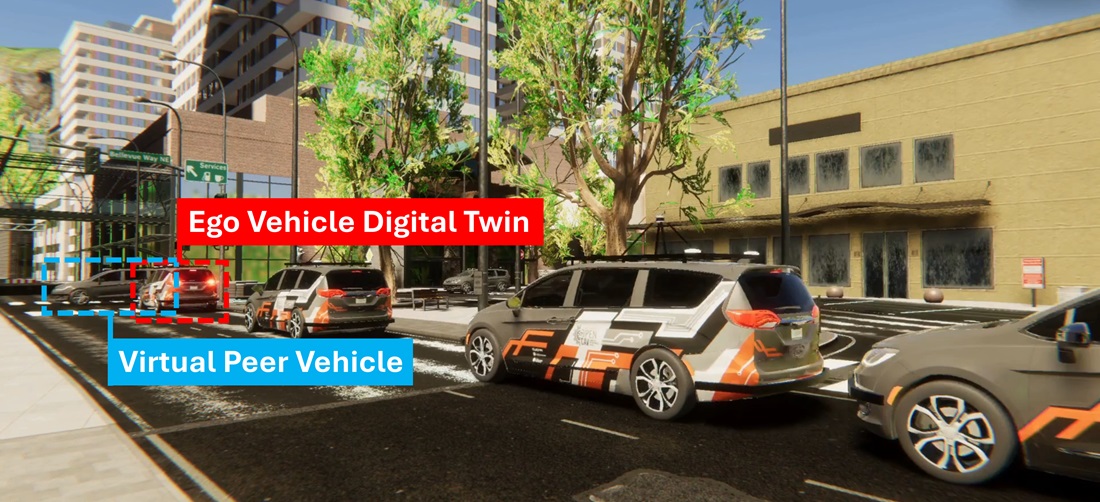}
         \caption{Ego vehicle digital twin with virtual traffic participant(s).}
         \label{fig5b}
     \end{subfigure}
     \caption{Freeze-frame sequence of the case study deployment. Video: \url{https://youtu.be/gYeeRgntkpA}}
    \label{fig5}
\end{figure*}

\subsection{Survey Instrument}
\label{Section: Survey Instrument}

We administered the immersive presence of users within the virtual environment of the proposed framework using the Bob G. Witmer Presence Questionnaire (PQ) \cite{PQ1998, PQ2005}. The PQ uses subjective questions such as \textit{``How much did the visual aspects of the environment involve you?''}, \textit{``How compelling was your sense of objects moving through space?''}, \textit{``How responsive was the environment to actions that you initiated (or performed)?''}, and \textit{``How natural was the mechanism which controlled movement through the environment?''} to measure user presence in virtual environments. Readers are suggested to refer \cite{PQ2005} for a complete list of questions. Each question is rated on a 7-point Likert scale, which determines the score of individual items. These items are then clustered based on principal component analysis to score the following 4 factors of immersive presence:

\begin{itemize}
    \item \textbf{Factor 1:} This factor measures \textit{involvement} of the users in virtual environments. PQ items 1, 2, 3, 4, 6, 7, 8, 10, 14, 17, 18, and 29 contribute to this factor.
    \item \textbf{Factor 2:} This factor measures \textit{sensory fidelity} of the framework in terms of the observation interface. PQ items 5, 11, 12, 13, 15, and 16 contribute to this factor.
    \item \textbf{Factor 3:} This factor measures \textit{adaptation/immersion} of the users in virtual environments. PQ items 9, 20, 21, 24, 25, 30, 31, and 32 contribute to this factor.
    \item \textbf{Factor 4:} This factor measures \textit{interface quality} of the framework in terms of negatively framed questions. PQ items 19, 22, and 23 contribute to this factor.
\end{itemize}

Particularly, we selected 21 questions out of the 29 item PQ (Version 3.0), and divided them into two sets. The first set comprised 11 questions (PQ04, PQ07, PQ08, PQ10, PQ14, PQ15, PQ16, PQ18, PQ20, PQ22, and PQ30), which primarily analyzed the observation interface of the framework. The second set comprised 10 questions (PQ01, PQ02, PQ03, PQ06, PQ09, PQ19, PQ21, PQ23, PQ24, and PQ31), which primarily analyzed the interaction interface of the framework. The omitted questions comprise auditory items (PQ05, PQ11, and PQ12), haptic items (PQ13, PQ17, and PQ29) and multi-modal sensing items (PQ25 and PQ32) in addition to the 3 questions (PQ26, PQ27 and PQ28) dropped from the 32-item PQ to the 29-item PQ in \cite{PQ2005}.


\section{User Study}
\label{Section: User Study}

\subsection{Participant Demographics}
\label{Section: Participant Demographics}

16 users (10 male and 6 female) of different age groups ($\textrm{min}=23$, $\textrm{max}=60$, $\mu=34.125$, $\sigma=13.271$ years old) willfully participated in our survey to analyze the user experience of various observation and interaction interfaces of the proposed framework. In terms of driving experience, 3 were proficient ($>10$ years), 3 were experienced ($6$-$10$ years), 4 were intermediate ($3$-$5$ years), 4 were beginners ($1$-$2$ years) and 2 had no driving experience. In terms of gaming experience, 5 were proficient ($>10$ years), 4 were experienced ($6$-$10$ years), 3 were intermediate ($3$-$5$ years), 2 were beginners ($1$-$2$ years) and 2 had never played a driving/racing video game. Finally, in terms of VR experience, 1 was proficient ($>5$ times), 1 was experienced ($3$-$4$ times), 9 were beginners ($1$-$2$ times), and 5 had no VR experience. 

\subsection{Design of Experiments}
\label{Section: US Design of Experiments}

We adopted 4 $\times$ 4 mixed factorial user studies to test 4 immersion levels and 4 interaction modalities with the proposed framework (refer Fig. \ref{fig2a}). Here, the order of exposing users to each configuration was the inter-subject independent variable, which was counterbalanced using a balanced Latin square. The intra-subject independent variable was the configuration itself. The dependent variable was the PQ score.

Configurations for observation interfaces:
\begin{itemize}
    \item \textbf{Single Monitor:} Users observed the virtual environment through a non-immersive HD (1920 $\times$ 1080) display.
    \item \textbf{Triple Monitor:} Users observed the virtual environment through 3 aligned non-immersive HD (5760 $\times$ 1080) displays.
    \item \textbf{Static HMD:} Users observed the virtual environment through a statically immersive HMD, wherein their head movements did not alter the point of view (POV).
    \item \textbf{Dynamic HMD:} Users observed the virtual environment through a dynamically immersive HMD, wherein their head movements changed the POV.
\end{itemize}

\begin{figure*}[t]
     \centering
     \includegraphics[width=\linewidth]{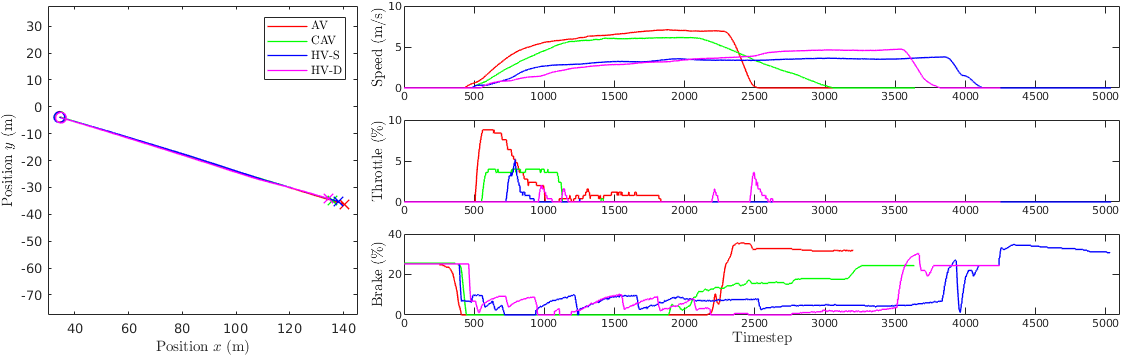}
     \caption{Ego vehicle behavior in autonomous (AV), connected autonomous (CAV), human-operated with static HMD (HV-S), and human-operated with dynamic HMD (HV-D) modes of operation.}
     \label{fig6}
\end{figure*}

Configurations for interaction interfaces:
\begin{itemize}
    \item \textbf{Keyboard:} Users controlled the ego vehicle using ASWD/arrow keys on a standard QWERTY keyboard.
    \item \textbf{Mouse:} Users controlled the ego vehicle by clicking and dragging the cursor across the screen.
    \item \textbf{Gamepad:} Users controlled the ego vehicle using two joysticks of a gamepad.
    \item \textbf{Driving Rig:} Users controlled the ego vehicle using a force feedback steering wheel ($\pm$450$^\circ$ range) and a set of pedals (accelerator and brake) while sitting in a driver's seat.
\end{itemize}

\subsection{Results and Discussion}
\label{Section: US Results and Discussion}

We analyzed the user experience responses for various observation (refer Fig. \ref{fig3}) and interaction (refer Fig. \ref{fig4}) interfaces. The overall trend suggests that users preferred the \{dynamic HMD + driving rig\} configuration the most; a more granular analysis follows.

In terms of observation interfaces, user ratings suggest a gradual increase in \textit{involvement} across the 4 modalities. \textit{Sensory fidelity} and user \textit{adaptation/immersion} for the first 3 configurations were rated almost similarly with a significantly higher rating for the fourth one. Finally, the \textit{interface quality} ratings suggest significantly higher discomfort/interference for the third configuration, followed by the first two, and least for the fourth one. It was observed that static HMD configuration confused some users (mostly those with prior VR experience) since they expected the HMD to change the POV of the rendered feed based on head movements.

In terms of interaction interfaces, user ratings again suggest a gradual increase in \textit{involvement} across the 4 modalities. User \textit{adaptation/immersion} for the first 3 configurations were rated almost similarly with a slightly higher rating for the fourth one. Finally, the \textit{interface quality} ratings suggest the highest discomfort/interference for the first two configurations, followed by the third one, and the least for the fourth one. It is worth mentioning that some users experienced occasional discomfort in locating certain interaction modalities (e.g. keys, cursor position, pedals) when wearing the HMD.


\section{Case Study}
\label{Section: Case Study}

\subsection{Driving Scenario}
\label{Section: Driving Scenario}

In order to evaluate the serviceability of the proposed framework, an edge-case urban driving scenario was devised. The scenario was laid out to jump-scare the ego vehicle and analyze its reactive response to the event. Particularly, the ego vehicle was supposed to keep driving straight unless it was unsafe/unethical to do so (e.g. dead end, traffic sign, traffic light, imminent collision, etc.). Down the road, a semi-autonomous virtual vehicle was set up to cut across the ego vehicle as it approached the first 4-way intersection, requiring the ego vehicle to execute a panic-breaking maneuver (refer Fig. \ref{fig5}) while ensuring safety in case of any failures.


\subsection{Design of Experiments}
\label{Section: CS Design of Experiments}


In addition to merely validating the aforementioned driving scenario, the case study aimed to explore the interaction dynamics between combinations of human-operated and autonomous vehicles. Particularly, the physical OpenCAV was operated on the Research Dr of CU-ICAR campus (refer Fig. \ref{fig2b}) in human-driven (dynamic HMD + gamepad), autonomous, and connected autonomous configurations, which ran in the loop with its digital twin and a secondary semi-autonomous virtual vehicle in an urban virtual environment.

It was ensured that all the experiments adhered to public road safety standards and involved a fall-back safety driver at all times.

\subsection{Results and Discussion}
\label{Section: CS Results and Discussion}

We comparatively analyzed the performance of the ego vehicle operated under different configurations viz. AV, CAV, HV-S, and HV-D (refer Fig. \ref{fig6}) in terms of its response to the jump scare scenario. Particularly, position, speed, throttle, and brake commands (indicative of reaction effort and time) were used as key performance indicators (KPIs).

Although all the configurations safely avoided collision with the virtual peer vehicle, their performance varied. The AV performed the most aggressive acceleration (2.35 m/s$^2$) and braking (-7.59 m/s$^2$) but traveled the farthest (110.21 m) before coming to a safe stop. This is likely because the peer vehicle was detected only after it was captured in the perception field of view. The CAV, on the other hand, was able to detect the presence of the peer vehicle ahead of time through V2V communication and therefore able to stop much earlier (105.49 m) and smoothly (-1.26 m/s$^2$). In terms of manual driving, it was noted that the observation interface greatly influenced the performance. The user with static HMD configuration performed worse than the CAV due to the same issue of being unable to observe the peer vehicle in time. Conversely, the user with dynamic HMD configuration could observe the peer vehicle approaching by turning their head and therefore performed the best out of all configurations in terms of stopping distance (103.86 m).


\section{Conclusion}
\label{Section: Conclusion}

This work presented an immersive digital twin framework aimed at safely exploring the intricate interactions between autonomous and non-autonomous traffic participants, essential for the widespread adoption of autonomous driving technology. By seamlessly integrating physical and digital assets through mixed-reality interfaces in real-time, we attempt to address the limitations inherent in both real-world and simulation-based methodologies, enabling safe exploration of diverse interaction scenarios. The findings from our user and case studies underscored the effectiveness and adaptability of the framework, establishing a groundwork for future research in autonomy-oriented digital twins and the coexistence of human drivers and autonomous systems.

Potential avenues for further research include developing innovative immersion techniques along the \textit{reality-virtuality continuum} including haptic and auditory feedback to expand the proposed framework. Additionally, involving more participants in the user study will improve its statistical significance. Lastly, realizing and validating complex driving scenarios with different scales and configurations of vehicles will further attest serviceability of the proposed framework.



\balance
\bibliographystyle{IEEEtran}
\bibliography{references}

\end{document}